# MGTUNet: An new UNet for colon nuclei instance segmentation and quantification

Liangrui Pan
College of Computer Science and
Electronic Engineering
HunanUniversity
Chang Sha, China
panlr@ hnu.edu.cn

Lian Wang
College of Computer Science and
Electronic Engineering
HunanUniversity
Chang Sha, China
lianwang@hnu.edu.cn

Zhichao Feng
Department of Radiology
Third Xiangya Hospital
Central South University
Chang Sha, China
fengzc2016@163.com

Zhujun Xu
College of Computer Science and
Electronic Engineering
HunanUniversity
Chang Sha, China
xzjcs1992@hotmail.com

Liwen Xu
College of Computer Science and
Electronic Engineering
Hunan University
Chang Sha, China
xuliwen @hnu.edu.cn

Shaoliang Peng*
College of Computer Science and
Electronic Engineering
Hunan University
Chang Sha, China
slpeng@hnu.edu.cn

*Abstract*—Colorectal cancer (CRC) is among the top three malignant tumor types in terms of morbidity and mortality. Histopathological images are the gold standard for diagnosing colon cancer. Cellular nuclei instance segmentation and classification, and nuclear component regression tasks can aid in the analysis of the tumor microenvironment in colon tissue. Traditional methods are still unable to handle both types of tasks end-to-end at the same time, and have poor prediction accuracy and high application costs. This paper proposes a new UNet model for handling nuclei based on the UNet framework, called MGTUNet, which uses Mish, Group normalization and transposed convolution layer to improve the segmentation model, and a ranger optimizer to adjust the SmoothL1Loss values. Secondly, it uses different channels to segment and classify different types of nucleus, ultimately completing the nuclei instance segmentation and classification task, and the nuclei component regression task simultaneously. Finally, we did extensive comparison experiments using eight segmentation models. By comparing the three evaluation metrics and the parameter sizes of the models, MGTUNet obtained 0.6254 on PQ, 0.6359 on mPQ, and 0.8695 on R2. Thus, the experiments demonstrated that MGTUNet is now a state-of-the-art method for quantifying histopathological images of colon cancer.

## I. INTRODUCTION

Colorectal cancer (CRC) is the most common cancer in the world and a major contributor to mortality, accounting for more than 9% of all cancer incidences. Although early CRC has a good prognosis with a five-year survival rate of over 90%, the detection rates are relatively low and most patients are diagnosed at an intermediate to late stage with a poor prognosis [1], [2]. The common method like fecal examination cannot meet the needs of accuracy screening and the supplementary methods are needed [3]. Usually, hematoxylin-eosin (HE)-stained tissue slides of CRC patients are used for detection and diagnosis. The histological slides highlight the nucleus and cytoplasm of tissue cells, which characterize quantitative information of the tumor. While subjective evaluation of histological slides by highly trained pathologists remains the gold standard for cancer diagnosis and staging, this remains labor-intensive and is limited by experience.

Further, cancer cells are highly heterogeneous. They are capable of causing varying degrees of host inflammatory response, angiogenesis and tumor necrosis. The spatial arrangement of these heterogeneous cell types has also been shown to correlate with cancer staging [4], [5]. Therefore, qualitative and quantitative analysis of different tumor types at the cellular level can help pathologists to better understand tumors and also to explore multiple options for treating cancer. The accurate localization of clinically relevant structures is the first step in pathological analysis. For example, segmentation of each nucleus yields a characterization of the nucleus morphology, the results of which can guide the prediction of cancer grade or survival analysis. Secondly, segmented cells reflecting nuclear morphological features need to be combined with the accurate classification of different cell types to quantify most cellular information on the whole-slide images (WSI). For example, the ratio of the number of the tumor to lymphocytes is often seen as a marker of prognosis in CRC. Histopathological images are therefore the gold standard for the diagnosis of cancer.

However, machine learning can train predictive models using the vast amount of image data information contained in billions of WSIs and has been widely applied in the field of computational pathology [6], [7]. Traditional machine learning algorithms require matrix eigenvalue decomposition, or singular value decomposition, and have a high time complexity to handle large amounts of data. Some algorithms need to calculate the distance between all sample points, which has a high spatio-temporal complexity. More importantly, many traditional learning algorithms are difficult to combine with GPU parallel computing and are therefore unable to handle large scale data. Deep learning is usually an unconstrained optimization problem, and the use of SGD can significantly improve training efficiency, benefiting from GPU parallel computing acceleration, so it can then handle large amounts of data. However, deep learning using an end-to-end approach to predicting WSIs can lead to models with poor interpretability, making it difficult for diagnoses to aid doctors or make patients credible. Explainable and fair AI algorithms have always been an aid to diagnosis that computational pathologists want.

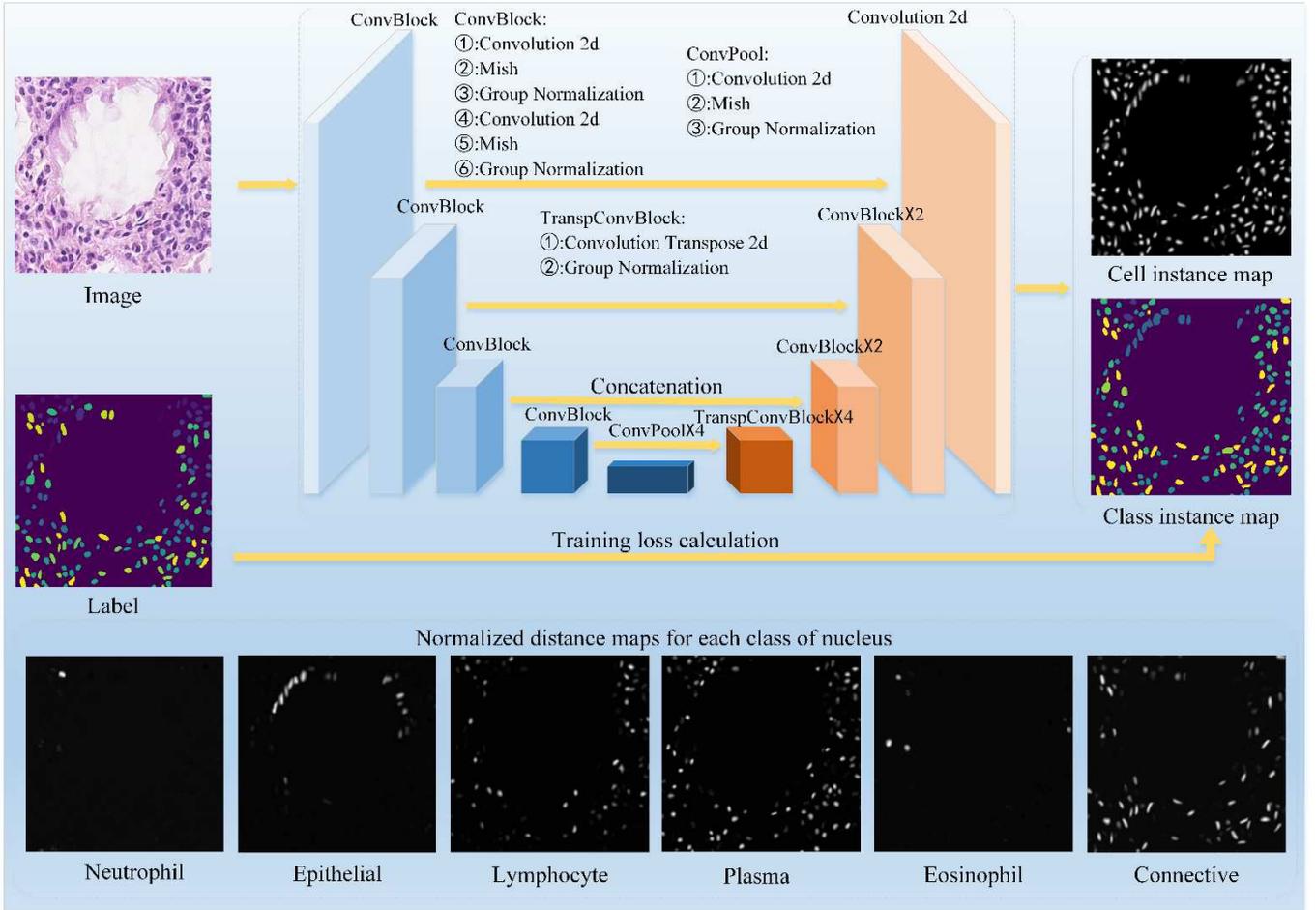

Fig. 1. Full flow chart of MGTUNet for segmenting histopathology images. (The training data includes images and instance ground truth. The prediction results include nuclei instance map and class instance map. Second, MGTUNet can also segment different types of nucleus.)

The original intention of UNet model design is to solve the problem of medical image segmentation. At present, it has been widely used in imaging data processing tasks combined with deep learning [8]. It uses a U-shaped network structure to obtain contextual information and location information, and reconstructs and predicts pathological images through feature extraction and up-sampling. The algorithm breakthrough of Transformer in recent years, combined with the UNet framework, has also been initially applied in the field of image segmentation [9]. However, for the segmentation task of the nucleus, this paper verifies that the feature extraction effect of some Transformers is not very significant. Secondly, the larger model and computing resource requirements lead to the inability to maximize the efficiency on the lizard dataset [10]. Therefore, this paper proposes a new nuclei processing UNet (MGTUNet) model to predict the type and location of the nucleus in pathology images. Our main contributions are as follows:

1. Based on the UNet framework, we propose to use Mish, Group normalization and transposed convolution layer to improve the segmentation model, and use the ranger optimizer to tune the SmoothL1Loss value.
2. MGTUNet uses different channels to segment and classify different types of nuclei, ultimately completing the nuclei instance segmentation and classification task, and the nuclei component regression task simultaneously.
3. We have done extensive comparison tests using the Cenet, PSPNet, UNet, Resnet50UNet, SegNet, R2UNet, Ternausnet and MGTUNet segmentation models and concluded that MGTUNet is the best method for quantifying histopathological images.

II. MATERIALS AND METHODS

A. Materials

The histological images of colon tissue used in the experiments were obtained from six different datasets which make up the lizard dataset [11]. At a magnification of 20×, the pathologists annotated the complete segmentation of the different nuclei, which contained six different nuclei labels, which were epithelial nucleus, connective tissue nucleus, lymphocytes, plasma nucleus, neutrophils and this nucleus accurately describe the colonic tumor microenvironment. To ensure that the lizard dataset contains a variety of different images of normal, inflammatory, dysplastic and cancerous diseases of the colon, the dataset adds the possibility of tagging unseen examples to generalize to other types. 4,981 patches of 256×256 size are provided with the lizard dataset. Each RGB image patch is associated with an instance segmentation map and a classification map, both of which are also 256×256 in size. Instance segmentation maps uniquely label each nucleus by containing values from 0

(background) to N (number of nuclei). The classification map provides the category of each pixel within the patch. Specifically, the nuclei classification map includes values from 0 (background) to C (number of classes). both the RGB image and the segmentation/classification map are stored as a single Numpy array. the RGB image array has dimensions of 4981×256×256×3 and is of type unsigned 8-bit integer, while the segmentation and classification map arrays have dimensions of 4981×256×256×2 and are of type unsigned signed 16-bit integer [12]. Here, the first channel is the instance segmentation map and the second channel is the classification map. 80% of the dataset is divided into a training set and the remaining 20% is divided into a test set.

*B. Methods*

In this paper, inspired by the UNet framework, we propose to present a completely new segmentation network for processing nuclei, called MGTUNet. Fig. 1 depicts the flowchart of MGTUNet segmentation. First, MGTUNet uses the first half of the network to extract features from histopathology image patches and corresponding nuclei labels. The second half of MGTUNet is used to generate predicted histopathology nuclei segmentation images and classification images, where each class of nucleus can be segmented and classified in a different channel.

MGTUNet consists of two parts, the encoder and the decoder. In particular, the encoder consists of four ConvBlock modules and four ConvPool modules. Each ConvBlock module is composed of two groups of Convolution 2d, Mish, Group Normalization (GN); each ConvPool module is composed of one group of Convolution 2d, Mish, GN. The decoder consists of four TranspConvBlock modules, two ConvBlock modules and a Convolution 2d layer. each TranspConvBlock module is composed of transposed convolution(TC) layer and GN. The prediction results generated by the decoder are output by the Convolution 2d layer.

Firstly, the histopathology image and labels are fed into MGTUNet's encoder before data augmentation such as horizontal/vertical flip, rotation, scaling, cropping, cropping, panning, etc. is performed. The Convolution layer in the ConvBlock module then performs feature extraction on the image. A convolution kernel of size 3×3 performs matrix computation on the image at a stride size of one. The positions around the matrix generated by the convolution operation that are zero are complemented by one. We process the feature matrix using the Mish activation function to incorporate non-linear factors to improve the expressiveness of the neural network, which can be expressed as:

$$f(x) = x * \tanh(\ln(1 + e^x)) \quad (1)$$

Where $x$ is the input image feature matrix and $f(x)$ is the output image feature matrix. Compared with Swish and ReLu, Mish has better capabilities of generalization and result optimization, and can improve the quality of results. The group normalization can also solve the Internal, Covariate and Shift problem with better results, and its computational procedure is as follows [13]:

$$y = \frac{x - E[x]}{\sqrt{Var[x] + \epsilon}} * \gamma + \beta \quad (2)$$

$x$ in the input channel is divided into several groups, and the mean and standard deviation of each group are calculated separately; $\gamma$ and $\beta$ are the sizes of the learnable per-channel affine transformation parameter vectors; the standard deviation is calculated by biased estimation. This layer uses statistics computed from the input data in both training and evaluation modes. After the pathological image is processed by four ConvBlock modules, its feature matrix will be preliminarily learned.

The ConvPool module will further process the pathological image features. The convolution nuclei in the convolution layer performs feature extraction with a stride of two, and the image features after the convolution operation are passed to Mish and GN for processing. So far, the work of the encoder part of MGTUNet has been completed.

The image feature matrix obtained by the encoder will be sent to the decoder. The TC layer in the TranspConvBlock module is a transposed convolution operation performed by a convolution kernel with a size of 2×2 with a stride size of two, which mainly plays the role of up-sampling. Unlike the convolution operation, it can only restore the feature matrix to the original size value, but it is different from the original. The process is: fill some values between the input feature map elements; fill some values around the input feature map; flip the convolution nuclei parameters up and down, left and right; do the classic convolution operation [14]. It can be expressed as:

$$o = \left\lfloor \frac{i - f + 2p}{s} \right\rfloor + 1 \quad (3)$$

where $i$ represents the size of the image input; $f$ represents the size of the convolution nuclei; $p$ represents the filled pixel value; $s$ represents the stride size; and o represents the size of the feature map, which is related to $i, f, p, s$ and also to the way the matrix is filled. The transposed convolution operation is followed by the transfer of the feature map to the GN. After feature extraction by the four ConvBlock modules, the feature map is finally output by the convolution layer.

The experiments use SmoothL1Loss to adjust the accuracy of the model. When the difference between the predicted value and the ground truth is small (the absolute value difference is less than 1), L2 Loss is actually used; when the difference is large, it is the translation of L1 Loss. SmoothL1Loss is actually a combination of L2Loss and L1Loss. It also has some advantages of L2 Loss and L1 Loss [15]. Its calculation can be expressed as:

$$loss(x, y) = \frac{1}{n}\sum_{i=1}^{n} \begin{cases} 0.5 * (y_i - f(x_i))^2, & if |y_i - f(x_i)| < 1 \\ |y_i - f(x_i)| - 0.5, & otherwise \end{cases} \quad (4)$$

$y$ denotes the true image of the model; $f(x)$ denotes the image predicted by the model. A simple cross-validation method allows us to obtain the final segmentation model. Some hyper-parameters must be mentioned in the experiments. The experiments use the ranger optimizer; the learning rate is set to 0.001, which will decrease as time grows; the batch size is set to 8 and the epoch to 50, and the training of the model stops early when the value of the loss remains constant. All models were trained on a GPU (NVIDIA V100) based on the PyTorch platform.

III. RESULT AND DISCUSSION

*A. Evaluation*

For the nuclei instance segmentation and classification tasks, we use a multi-class panoramic quality (PQ) to reflect the performance of the model, and for each type t, $PQ_t$ is defined as:

$$PQ_t = \frac{|TP_t|}{|TP_t| + \frac{1}{2}|FP_t| + \frac{1}{2}|FN_t|} \times \frac{\sum_{(x_t, y_t) \in TP} IoU(x_t, y_t)}{|TP_t|} \quad (5)$$

where $x$ denotes a ground truth (GT) instance, $y$ denotes the predicted instance, and IoU denotes intersection over union. $IoU(x, y) > 0.5$ will uniquely match $x$ and $y$. Each type of t can be split into matched pairs (TP), unmatched GT instances (FN) and unmatched predicted instances (FP). We define the multiclass PQ as the evaluation metric for the experiment, which has an all-class mean of:

$$mPQ = \frac{1}{T} \sum_t PQ_t \quad (6)$$

For the task of nuclei component regression, the experiments use a multi-class coefficient of determination to determine the correlation between predicted and true counts. To do this, the statistics for each class are computed independently, and the results are then averaged. Therefore, for each nuclei class t, its correlation can be defined as:

$$R_t^2 = 1 - \frac{RSS_t}{TSS_t} \quad (7)$$

where $RSS$ denotes the residual sum of squares and $TSS$ denotes the total sum of squares.

*B. Results*

Due to the small sample size of the experiment, we used data augmentation to enrich the training set. The experiments use two datasets generated based on the lizard dataset, the original one and the up-sampled one. First, we trained the Cenet, PSPNet, UNet, Resnet50UNet, SegNet, R2UNet, Ternausnet and MGTUNet segmentation models using 80% of the data from the up-sampled dataset. The remaining 20% of the data was used to test the models. To compare the performance of the models for nuclei instance segmentation and nuclei component regression, we counted $PQ_t$, mPQ and R2 in TABLE 1. the values of PQ and mPQ for UNet reached 0.6035 and 0.6409, 0.032 and 0.0112 higher than for MGTUNet, respectively. the value of R2 for MGTUNet reached 0.7881, 0.0008 larger than for UNet. However, the performance of the other UNet-based models was average.

TABLE I.  PERFORMANCE COMPARISON OF MODELS TRAINED ON THE UPSAMPLED LIZARD DATASET.

|  | PQ | mPQ | R2 |
|---|---|---|---|
| Cenet | 0.5079 | 0.5233 | 0.2009 |
| PSPNet | 0.5702 | 0.604 | 0.7088 |
| UNet | **0.6035** | **0.6409** | 0.7872 |
| Resnet50UNet | 0.5989 | 0.6322 | 0.7396 |
| SegNet | 0.4841 | 0.4665 | 0.1441 |
| R2UNet | 0.5202 | 0.5385 | 1.1389 |
| Ternausnet | 0.5534 | 0.6116 | 0.4789 |
| MGTUNet | 0.6003 | 0.6297 | **0.7881** |

For the lizard dataset trained on the no up-sampling dataset, we still used 80% of the data to train the Cenet, PSPNet, UNet, Resnet50UNet, SegNet, R2UNet, Ternausnet and MGTUNet segmentation models. The remaining 20% of the data was used to test the models. Based on the evaluation metrics PQ, mPQ and R2, we tabulated the test results for the eight models in TABLE 2. We found that UNet still had values for PQ and mPQ that were 0.075 and 0.094 larger than MGTUNet. however, MGTUNet had a value for R2 of 0.8695, which was 0.0011 larger than UNet.

TABLE II.  PERFORMANCE COMPARISON OF MODELS TRAINED ON THE NO UPSAMPLING LIZARD DATASET.

|  | PQ | mPQ | R2 |
|---|---|---|---|
| Cenet | 0.489 | 0.4947 | 0.25891 |
| PSPNet | 0.5357 | 0.5563 | 0.7637 |
| UNet | **0.6329** | **0.6445** | 0.8684 |
| Resnet50UNet | 0.6117 | 0.6251 | 0.8548 |
| SegNet | 0.4715 | 0.4469 | 0.3063 |
| R2UNet | 0.5099 | 0.5346 | -3.8478 |
| Ternausnet | 0.5663 | 0.6043 | 0.4942 |
| MGTUNet | 0.6254 | 0.6359 | **0.8695** |

Given that the size of the model affects computational resources, training time and application costs, the parameter size of the model should not be ignored. In order to fairly compare the performance of the models, we used the parameter sizes of the models for comparison. The parameter size of UNet is 143.69 MB; the parameter size of MGTUNet is 130.39 MB. therefore, using PQ, mPQ, R2 and the parameter size of the model, we can determine that MGTUNet is the nuclei instance segmentation and classification task with the nuclei component regression task as a standard method.

Furthermore, according to the predicted nuclei types, our model quantified the cell count of six different cell types in the tumor microenvironment in Fig. 2. The computational inference of cell count from WSI may effectively supplement current methods in revealing spatial tumor-infiltrating lymphocytes (TIL) patterns and cellular compositions. We expect this histopathological framework to improve TME predictions and even construct promising biomarkers for tumor immunotherapy in the future.

*C. Discussion*

In our experiments, we also used transformer-based UNet frameworks such as SwinUNet, Transfuse and TransUNet models for the nuclei instance segmentation and classification with nuclei component regression tasks [16]–[18]. The training time of these models is approximately four times longer than that of the UNet model, and the prediction performance of these models is lower than UNet. Therefore, we did not use the transformer-based UNet framework to design new models. Second, training the model on the up-sampled dataset was generally not as effective as training on the original dataset. It is possible that some of the same samples were present in the up-sampled dataset, resulting in no improvement in the model training. Secondly, the large number of samples may have caused the model training to be biased towards identifying a particular type of nucleus.

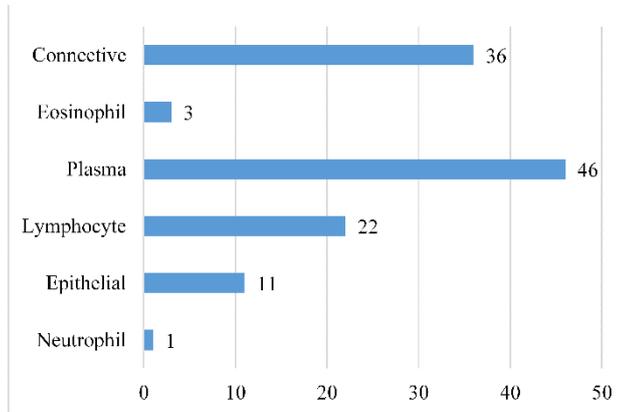

Fig. 2. Quantified cell count in tumor microenvironment of one sample by the model.

## IV. CONCLUSION

This paper proposes a new colon nuclei processing model based on the UNet framework, named MGTUNet, which uses Mish, Group normalization and transposed convolution layer to improve the segmentation model, and a ranger optimizer to adjust the SmoothL1Loss value. Secondly, it uses different channels to segment and classify different types of nucleus, ultimately completing the segmentation task and the classification task simultaneously. Finally, we did extensive comparison experiments using the Cenet, PSPNet, UNet, Resnet50UNet, SegNet, R2UNet, Ternausnet and MGTUNet segmentation models. The parameter sizes of the models were compared by three evaluation metrics, PQ, mPQ and R2, and MGTUNet obtained 0.6254, 0.6359 and 0.8695 for PQ, mPQ and R2 respectively. its model had the smallest parameter of all models at 130.39 MB. in summary, the experiments concluded that MGTUNet is the best method for quantifying colon histopathological images.

## ACKNOWLEDGEMENTS

This work was supported by NSFC Grants U19A2067; Science Foundation for Distinguished Young Scholars of Hunan Province (2020JJ2009); National Key R&D Program of China 2017YFB0202602, 2018YFC0910405, 2017YFC1311003, 2016YFC1302500; Science Foundation of Changsha Z202069420652, kq2004010; JZ20195242029, JH20199142034; The Funds of State Key Laboratory of Chemo/Biosensing and Chemometrics and Peng Cheng Lab.